\documentclass{article}

\usepackage{microtype}
\usepackage{graphicx}
\usepackage{subfigure}
\usepackage{booktabs} 
\usepackage{rotating}
\usepackage{multirow}
\usepackage{amsmath}
\usepackage{amssymb}
\usepackage{algorithm2e}
\usepackage{hyperref}
\usepackage[accepted]{icml2020}

\begin{document}

\twocolumn[
\icmltitle{Contrastive Multi-View Representation Learning on Graphs}
\icmlsetsymbol{equal}{*}

\begin{icmlauthorlist}
\icmlauthor{Kaveh Hassani}{adsk}
\icmlauthor{Amir Hosein Khasahmadi}{adsk,vec}
\end{icmlauthorlist}

\icmlaffiliation{adsk}{Autodesk AI Lab, Toronto, Canada}
\icmlaffiliation{vec}{Vector Institute, Toronto, Canada}

\icmlcorrespondingauthor{Kaveh Hassani}{kaveh.hassani@autodesk.com}
\icmlcorrespondingauthor{Amir Hosein Khasahmadi}{amir.khasahmadi@autodesk.com}

\icmlkeywords{Self-supervised Learning, Graph Representation Learning, Graph Neural Network}
\vskip 0.3in
]
\printAffiliationsAndNotice{}

\begin{abstract}
We introduce a self-supervised approach for learning node and graph level representations by contrasting structural views of graphs. We show that unlike visual 
representation learning, increasing the number of views to more than two or contrasting multi-scale encodings do not improve performance, and the best 
performance is achieved by contrasting encodings from first-order neighbors and a graph diffusion. We achieve new state-of-the-art results in self-supervised learning on 
8 out of 8 node and graph classification benchmarks under the linear evaluation protocol. For example, on Cora (node) and Reddit-Binary (graph) classification 
benchmarks, we achieve 86.8\% and 84.5\% accuracy, which are 5.5\% and 2.4\% relative improvements over previous state-of-the-art. When compared to 
supervised baselines, our approach outperforms them in 4 out of 8 benchmarks.
\end{abstract}
\section{Introduction}
Graph neural networks (GNN) \cite{li_2015_iclr, gilmer_2017_icml, kipf_2017_iclr, velickovic_2018_iclr, xu_2019_iclr} reconcile the expressive power of graphs in
modeling interactions with unparalleled capacity of deep models in learning representations. They process variable-size permutation-invariant graphs and learn 
low-dimensional representations through an iterative process of transferring, transforming, and aggregating the representations from topological neighbors. Each 
iteration expands the \textit{receptive field} by one-hop and after $k$ iterations the nodes within $k$-hops influence one another \cite{Khasahmadi_2020_iclr}. GNNs are
applied to data with arbitrary topology such as point clouds \cite{hassani_2019_iccv}, meshes \cite{wang_2018_eccv}, robot designs \cite{wang_2018_iclr}, physical processes \cite{sanchez_2018_icml}, social networks \cite{kipf_2017_iclr}, molecules \cite{duvenaud_2015_nips}, and knowledge graphs \cite{vivona_2019_nips}. For an
overview see \cite{zhang_2020_kde, wu_2020_nnls}.

GNNs mostly require task-dependent labels to learn rich representations. Nevertheless, annotating graphs is challenging compared to more common modalities 
such as video, image, text, and audio. This is partly because graphs are usually used to represent concepts in specialized domains, e.g., biology. Also, labeling 
graphs procedurally using domain knowledge is costly \cite{Sun_2020_iclr}. To address this, unsupervised approaches such as reconstruction based methods 
\cite{kipf_2016_arxiv} and contrastive methods \cite{li_2019_icml} are coupled with GNNs to allow them learn representations without relying on supervisory data. 
The learned representations are then transferred to a priori unknown down-stream tasks. Recent works on contrastive learning by maximizing mutual information 
(MI) between node and graph representations have achieved state-of-the-art results on both node classification \cite{velickovic_2019_iclr} and graph classification 
\cite{Sun_2020_iclr} tasks. Nonetheless, these methods require specialized encoders to learn graph or node level representations.

Recent advances in multi-view visual representation learning \cite{tian_2019_arxiv, bachman_2019_nips, chen_2020_arxiv}, in which composition of data 
augmentations is used to generate multiple views of a same image for contrastive learning, has achieved state-of-the-art results on image classification 
benchmarks surpassing supervised baselines. However, it is not clear how to apply these techniques to data represented as graphs. To address this, we introduce a 
self-supervised approach to train graph encoders by maximizing MI between representations encoded from different structural views of graphs. We show that our 
approach outperforms previous self-supervised models with significant margin on both node and graph classification tasks without requiring specialized 
architectures. We also show that when compared to supervised baselines, it performs on par with or better than strong baselines on some benchmarks. 

To further improve contrastive representation learning on node and graph classification tasks, we systematically study the major components of our 
framework and surprisingly show that unlike visual contrastive learning: (1) increasing the number of views, i.e., augmentations, to more than two views does not improve the performance and the best performance is achieved by contrasting encodings from first-order neighbors and a general graph diffusion, (2) contrasting node and graph encodings across views achieves better results on both tasks compared to contrasting graph-graph or multi-scale encodings, (3) a simple graph readout layer achieves better performance on both tasks compared to hierarchical graph pooling methods such as differentiable pooling (DiffPool) 
\cite{ying_2018_nips}, and (4) applying  regularization (except early-stopping) or normalization layers has a negative effect on the performance.

Using these findings, we achieve new state-of-the-art in self-supervised learning on 8 out of 8 node and graph classification benchmarks under the linear evaluation protocol. For example, on Cora node classification benchmark, our approach achieves 86.8\% accuracy, which is a 5.5\% relative improvement over 
previous state-of-the-art \cite{velickovic_2019_iclr}, and on Reddit-Binary graph classification benchmark, it achieves 84.5\% accuracy, i.e., a 2.4\% relative 
improvement over previous state-of-the-art \cite{Sun_2020_iclr}. When compared to supervised baselines, our approach performs on par with or better than strong
supervised baselines, e.g., graph isomorphism network (GIN) \cite{xu_2019_iclr} and graph attention network (GAT) \cite{velickovic_2018_iclr}, on 4 out of 8 
benchmarks. As an instance, on Cora (node) and IMDB-Binary (graph) classification benchmarks, we observe 4.5\% and 5.3\% relative improvements over GAT, 
respectively.
\section{Related Work}
\subsection{Unsupervised Representation Learning on Graphs}
\textbf{Random walks} \cite{perozzi_2014_kdd, tang_2015_www, grover_2016_kdd, hamilton_2017_nips} flatten graphs into sequences by taking random walks across 
nodes and use language models to learn node representations. They are shown to over-emphasize proximity information at the expense of structural information 
\cite{velickovic_2019_iclr,ribeiro_2017_kdd}. Also, they are limited to transductive settings and cannot use node features \cite{you_2019_icml}.\\
\textbf{Graph kernels} \cite{borgwardt_2005_icdm, shervashidze_2009_ais, shervashidze_2011_jmlr, yanardag_2015_kdd, kondor_2016_nips, kriege_2016_nips} decompose graphs into sub-structures and use kernel functions to measure graph similarity between them. Nevertheless, they require non-trivial task of devising
similarity measures between substructures. \\
\textbf{Graph autoencoders (GAE)} \cite{kipf_2016_arxiv, duran_2017_nips, wang_2017_cikm, pan_2018_ijcai, park_2019_iccv} train encoders that impose the topological 
closeness of nodes in the graph structure on the latent space by predicting the first-order neighbors. GAEs over-emphasize proximity information 
\cite{velickovic_2019_iclr} and suffer from unstructured predictions \cite{tian_2019_arxiv}. \\
\textbf{Contrastive methods} \cite{li_2019_icml, velickovic_2019_iclr, Sun_2020_iclr} measure the loss in latent space by contrasting samples from a distribution that
contains dependencies of interest and the distribution that does not. These methods are the current state-of-the-art in unsupervised node and graph classification 
tasks. Deep graph Infomax (DGI) \cite{velickovic_2019_iclr} extends deep InfoMax \cite{hjelm_2019_iclr} to graphs and achieves state-of-the-art results in node 
classification benchmarks by learning node representations through contrasting node and graph encodings. InfoGraph \cite{Sun_2020_iclr}, on the other hand, 
extends deep InfoMax to learn graph-level representations and outperforms previous models on unsupervised graph classification tasks. Although these two methods use the same contrastive learning approach, 
they utilize specialized encoders.
\subsection{Graph Diffusion Networks}
Graph diffusion networks (GDN) reconcile spatial message passing and generalized graph diffusion \cite{klicpera_2019_nips} where diffusion as a denoising filter 
allows messages to pass through higher-order neighborhoods. GDNs can be categorized to early- and late- fusion models based on the stage the diffusion is used. 
Early-fusion models \cite{xu_2019_ijcai, jiang_2019_cvpr} use graph diffusion to decide the neighbors, e.g., graph diffusion convolution (GDC) replaces adjacency 
matrix in graph convolution with a sparsified diffusion matrix \cite{klicpera_2019_nips}, whereas, late-fusion models \cite{tsitsulin_2018_www, klicpera_2019_iclr} 
project the node features into a latent space and then propagate the learned representation based on a diffusion.
\subsection{Learning by Mutual Information Maximization}
\label{contrastive overview}
InfoMax principle \cite{linsker_1988_computer} encourages an encoder to learn representations that maximizes the MI between the input and the learned 
representation. Recently, a few self-supervised models inspired by this principle are proposed which estimate the lower bound of the InfoMax objective, e.g., 
using noise contrastive estimation \cite{gutmann_2010_aistat}, across representations. Contrastive predictive coding (CPC) \cite{oord_2018_arxiv} contrasts a 
summary of ordered local features to predict a local feature in the future whereas deep InfoMax (DIM) \cite{hjelm_2019_iclr} simultaneously contrasts a single 
summary feature, i.e., global feature, with all local features. Contrastive multiview coding (CMC) \cite{tian_2019_arxiv}, augmented multi-scale DIM (AMDIM) 
\cite{bachman_2019_nips}, and SimCLR \cite{chen_2020_arxiv} extend the InfoMax principle to multiple views and maximize the MI across views generated by 
composition of data augmentations. Nevertheless, it is shown that success of these models cannot only be attributed to the properties of MI
alone, and the choice of encoder and MI estimators have a significant impact on the performance \cite{tschannen_2020_iclr}.
\begin{figure*}[ht] 
\label{figarch}
\vskip 0.2in
\begin{center}
\centerline{\includegraphics[width=170mm]{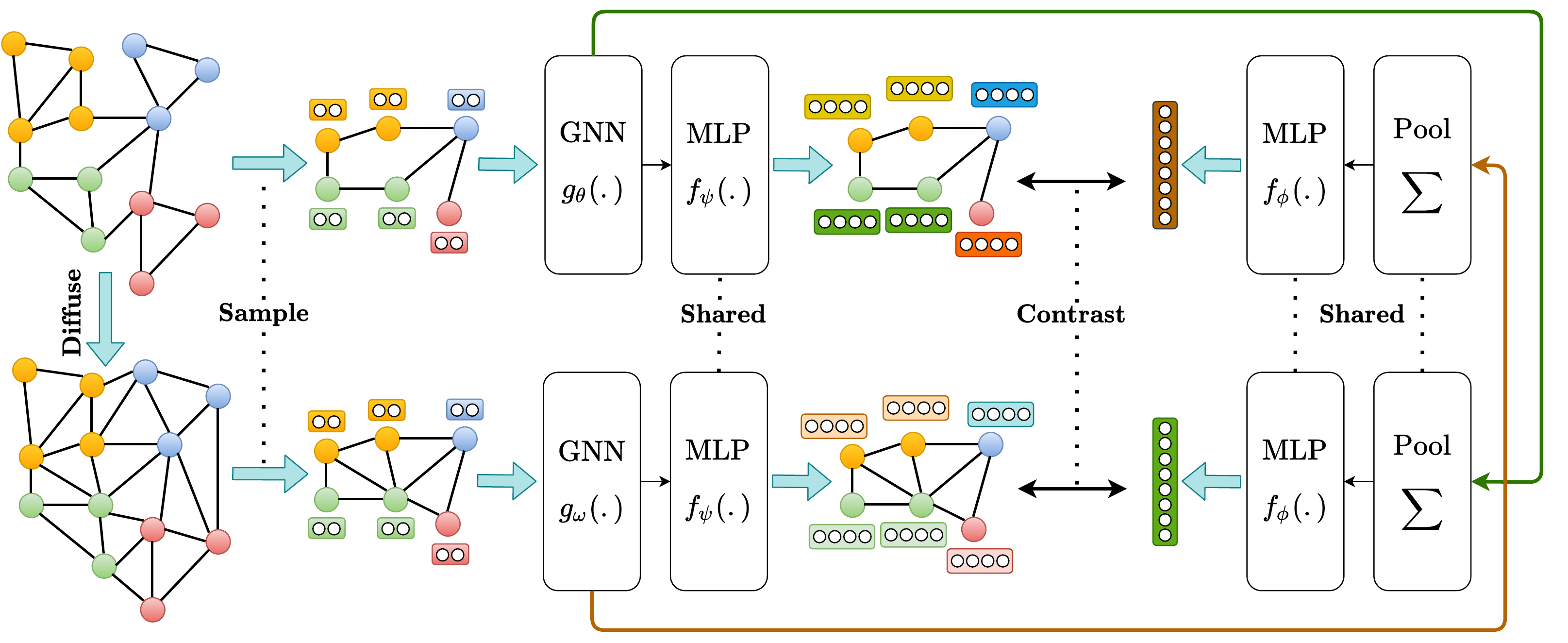}}
\caption{The proposed model for contrastive multi-view representation learning on both node and graph levels. A graph diffusion is used to generate an 
additional structural view of a sample graph which along with a regular view are sub-sampled and fed to two dedicated GNNs followed by a shared MLP to learn 
node representations. The learned features are then fed to a graph pooling layer followed by a shared MLP to learn graph representations. A discriminator 
contrasts node representations from one view with graph representation of another view and vice versa, and scores the agreement between representations which 
is used as the training signal.}
\end{center}
\vskip -0.2in
\end{figure*}
\section{Method}
Inspired by recent advances in multi-view contrastive learning for visual representation learning, our approach learns node and graph representations by maximizing MI between node representations of one view and graph representation of another view and vice versa which achieves better results compared to 
contrasting global or multi-scale encodings on both node and graph classification tasks (see section \ref{sec:ablation}). As shown in Figure 1, our method 
consists of the following components: 
\begin{itemize}
\item An augmentation mechanism that transforms a sample graph into a correlated view of the same graph. We only apply the augmentation to the structure of the graphs and not the initial node features. This is followed by a sampler that sub-samples identical nodes from both views, i.e., similar to cropping in visual domain.
\item Two dedicated GNNs, i.e., graph encoders, one for each view, followed by a shared MLP, i.e., projection head, to learn node representations for both views.
\item A graph pooling layer, i.e., readout function, followed by a shared MLP, i.e., projection head, to learn graph representations for both views.
\item A discriminator that contrasts node representations from one view with graph representation from another view and vice versa, and scores the agreement between them.
\end{itemize}
\subsection{Augmentations}
\label{augmentation}
Recent works in self-supervised visual representation learning suggest that contrasting congruent and incongruent views of images allows encoders to learn rich
representations \cite{tian_2019_arxiv, bachman_2019_nips}. Unlike images in which views are generated by standard augmentations, e.g., cropping, rotating, 
distorting colors, etc., defining views on graphs is not a trivial task. We can consider two types of augmentations on graphs: (1) feature-space augmentations 
operating on initial node features, e.g., masking or adding Gaussian noise, and (2) structure-space augmentations and corruptions operating on graph structure by
adding or removing connectivities, sub-sampling, or generating global views using shortest distances or diffusion matrices. The former augmentation can be 
problematic as many benchmarks do not carry initial node features. Moreover, we observed that masking or adding noise on either spaces degrades the performance. Hence, we opted for generating a global view followed by sub-sampling.

We empirically show that in most cases the best results are achieved by transforming an adjacency matrix to a diffusion matrix and treating the two matrices as 
two congruent views of the same graph's structure (see section \ref{sec:ablation}). We speculate that because adjacency and diffusion matrices provide local and global views of a graph structure, respectively, maximizing agreement between representation learned from these two views allows the model to simultaneously 
encode rich local and global information.

Diffusion is formulated as Eq. (\ref{eq:ggd}) where $\textbf{T} \in \mathbb{R}^{n\times n}$ is the generalized transition matrix and $\Theta$ is the weighting coefficient which determines the ratio of 
global-local information. Imposing $\sum_{k=0}^{\infty} \theta_k=1$, $\theta_k \in [0,1]$, and $\lambda_i \in [0,1]$ where $\lambda_i$ are eigenvalues of \textbf{T}, guarantees convergence. Diffusion is computed once 
using fast approximation and sparsification methods \cite{klicpera_2019_nips}.
\begin{equation}
    \label{eq:ggd}
	\textbf{S}=\sum_{k=0}^{\infty}{\Theta_k \textbf{T}^k} \in \mathbb{R}^{n\times n}
\end{equation}
Given an adjacency matrix $\textbf{A}\in \mathbb{R}^{n\times n}$ and a diagonal degree matrix $\textbf{D}\in \mathbb{R}^{n\times n}$, Personalized PageRank (PPR) \cite{page_1999_stanford} and heat kernel 
\cite{kondor_2002_icml}, i.e., two instantiations of the generalized graph diffusion, are defined by setting $\mathbf{T}=\textbf{AD}^{-1}$, and $\theta_k=\alpha(1-\alpha)^k$ and $\theta_k= e^{-t}t^{k}/{k!}$, respectively, where $\alpha$ denotes teleport probability in a random walk and $t$ is diffusion time \cite{klicpera_2019_nips}. Closed-form solutions to heat and PPR 
diffusion are formulated in Eq. (\ref{eq:heat}) and (\ref{eq:ppr}), respectively.
\begin{equation}
    \label{eq:heat}
	\textbf{S}^{\text{heat}}=\text{exp}\left({t\textbf{A}\textbf{D}^{-1}-t}\right) 
\end{equation}
\begin{equation}
    \label{eq:ppr}
\begin{aligned}
	\textbf{S}^{\text{PPR}}=\alpha\left(\textbf{I}_n-(1-\alpha) \textbf{D}^{-1/2}\textbf{A}\textbf{D}^{-1/2}\right)^{-1}
    \end{aligned}
\end{equation}
For sub-sampling, we randomly sample nodes and their edges from one view and select the exact nodes and edges from the other view. This procedure allows our 
approach to be applied to inductive tasks with graphs that do not fit into the GPU memory and also to transductive tasks by considering sub-samples as 
independent graphs.
\subsection{Encoders}
\label{encoder}
Our framework allows various choices of the network architecture without any constraints. We opt for simplicity and adopt the commonly used graph 
convolution network (GCN) \cite{kipf_2017_iclr} as our base graph encoders. As shown in Figure 1, we use a dedicated graph encoder for each view, i.e., 
$g_\theta(.), g_\omega(.): \mathbb{R}^{n \times d_x} \times \mathbb{R}^{n \times n} \longmapsto \mathbb{R}^{n \times d_h}$. We consider adjacency and diffusion matrices as two congruent structural views and define GCN layers as 
$\sigma(\mathbf{\tilde{A}}\mathbf{X}\mathbf{\Theta })$ and $\sigma\big(\mathbf{S}\mathbf{X}\mathbf{\Theta}\big)$to learn two sets of node representations each corresponding to one of the views, respectively. $\mathbf{\tilde{A}}=\mathbf{\hat{D}}^{-1/2} \mathbf{\hat{A}}\mathbf{\hat{D}}^{-1/2}\in \mathbb{R}^{n \times n}$ is symmetrically normalized adjacency matrix, $\mathbf{\hat{D}}\in \mathbb{R}^{n \times n}$ is the degree matrix of $\mathbf{\hat{A}} = \mathbf{A} + \mathbf{I_N}$ where $\mathbf{I_N}$ is the identity matrix, $\textbf{S}\in \mathbb{R}^{n \times n}$ is diffusion matrix, 
$\textbf{X}\in \mathbb{R}^{n \times d_x}$ is the initial node features, $\mathbf{\Theta}\in \mathbb{R}^{d_x \times d_h}$ is network parameters, and $\sigma$ is a parametric ReLU (PReLU) non-linearity \cite{he_2015_iccv}. The learned representations are then fed 
into a shared projection head $f_\psi(.): \mathbb{R}^{n \times d_h}\longmapsto \mathbb{R}^{n \times d_h}$ which is an MLP with two hidden layers and PReLU non-linearity. This results in two sets of node 
representations $\mathbf{H}^\alpha$, $\mathbf{H}^\beta \in \mathbb{R}^{n \times d_h}$ corresponding to two congruent views of a same graph.

For each view, we aggregate the node representations learned by GNNs, i.e., before the projection head, into a graph representation using a graph pooling 
(readout) function $
\mathcal{P}(.) : \mathbb{R}^{n \times d_h} \longmapsto \mathbb{R}^{d_h}$. We use a readout function similar to jumping knowledge network (JK-Net) \cite{Xu_2018_icml} where we concatenate 
the summation of the node representations in each GCN layer and then feed them to a single layer feed-forward network to have a consistent dimension size 
between node and graph representations:
\begin{equation}
\vec{h}_g=\sigma\left({\operatorname*{\parallel}_{l=1}^{L} \left[ \sum_{i=1}^n \vec{h}_i^{(l)}\right]}\mathbf{W}\right) \in \mathbb{R}^{h_d}
\end{equation}
where $\vec{h}_i^{(l)}$ is the latent representation of node $i$ in layer $l$, $||$ is the concatenation operator, $L$ is the number of GCN layers, $\mathbf{W}\in\mathbb{R}^{(L\times d_h) \times d_h}$ is the network 
parameters, and $\sigma$ is a PReLU non-linearity. In section \ref{sec:ablation}, we show that this pooling function achieves better results compared to more complicated 
graph pooling methods such as DiffPool \cite{ying_2018_nips}. Applying the readout function to node representations results in two graph representations, each 
associated with one of the views. The representations are then fed into a shared projection head $f_\phi(.): \mathbb{R}^{d_h}\longmapsto \mathbb{R}^{d_h}$, which is an MLP with two hidden layers and 
PReLU non-linearity, resulting in the final graph representations: $\vec{h}_g^\alpha, \vec{h}_g^\beta \in \mathbb{R}^{d_h}$.

At inference time, we aggregate the representations from both views in both node and graph levels by summing them up: $\vec{h} = \vec{h}_g^\alpha +\vec{h}_g^\beta \in \mathbb{R}^n$ and 
$\mathbf{H} = \mathbf{H}^\alpha + \mathbf{H}^\beta \in \mathbb{R}^{n \times d_h}$ and return them as graph and node representations, respectively, for the down-stream tasks.
\subsection{Training}
In order to train the encoders end-to-end and learn rich node and graph level representations that are agnostic to down-stream tasks, we utilize the deep InfoMax
\cite{hjelm_2019_iclr} approach and maximize the MI between two views by contrasting node representations of one view with graph representation of the other 
view and vice versa. We empirically show that this approach consistently outperforms contrasting graph-graph or multi-scale encodings on both node and graph 
classifications benchmarks (see section \ref{sec:ablation}). We define the objective as follows:
\begin{equation}
\max_{\theta, \omega, \phi, \psi} \frac{1}{|\mathcal{G}|} \sum_{g \in \mathcal{G}} \left[ \frac{1}{|g|} \sum_{i=1}^{|g|} \left[ \text{MI}\left( \vec{h}_i^\alpha, \vec{h}_g^\beta \right) + \text{MI}\left( \vec{h}_i^\beta, \vec{h}_g^\alpha \right)\right] \right]
\end{equation}
where ${\theta, \omega, \phi, \psi}$ are graph encoder and projection head parameters, ${|\mathcal{G}|}$ is the number of graphs in train set or number of sub-sampled graphs in transductive setting, ${|g|}$ is the number of nodes in graph $g$, and $\vec{h}_i^\alpha, \vec{h}_g^\beta$ are representations of node $i$ and graph $g$ encoded from views $\alpha, \beta$, respectively. 

MI is modeled as discriminator $\mathcal{D}(., .):\mathbb{R}^{d_h} \times \mathbb{R}^{d_h} \longmapsto \mathbb{R}$ that takes in a node representation from one view and a graph representation from another view, and scores the agreement between them. We simply implement the discriminator as the dot product between two representations: $\mathcal{D}(\vec{h}_n, \vec{h}_g)=\vec{h}_n.\vec{h}_g^T$. We observed a 
slight improvement in node classification benchmarks when discriminator and projection heads are integrated into a bilinear layer. In order to decide the MI 
estimator, we investigate four estimators and chose the best one for each benchmark (see section \ref{sec:ablation}). 

We provide the positive samples from joint distribution $ x_p \sim p\left(\left[\mathbf{X},\tau_\alpha\left(\mathbf{A}\right)\right], \left[\mathbf{X},\tau_\beta\left(\mathbf{A}\right)\right]\right)$ and the negative samples from the product of marginals
$ x_n \sim p\left(\left[\mathbf{X},\tau_\alpha(\mathbf{A}\right)\right])p(\left[\mathbf{X},\tau_\beta\left(\mathbf{A}\right)\right])$. To generate negative samples in transductive tasks, we randomly shuffle the features \cite{velickovic_2019_iclr}. Finally, we 
optimize the model parameters with respect to the objective using mini-batch stochastic gradient descent. Assuming a set of training graphs $\mathcal{G}$ where a sample
graph $g=(\textbf{A}, \textbf{X})\in \mathcal{G} $ consists of an adjacency matrix $\textbf{A}\in\lbrace 0, 1 \rbrace^ {n\times n}$ and initial node features $\textbf{X}\in \mathbb{R}^{n \times d_x}$, our proposed multi-view representation learning 
algorithm is summarized as follows:

\begin{algorithm}
\SetAlgoLined \DontPrintSemicolon
\KwIn{Augmentations $\tau_\alpha$ and $\tau_\beta$, sampler $\Gamma$, pooling $\mathcal{P}$, discriminator $\mathcal{D}$, loss $\mathcal{L}$, encoders $g_\theta$, $g_\omega$, $f_\psi$, $f_\phi$, and training graphs $\{\mathcal{G}|g=(\mathbf{X},\mathbf{A}) \in \mathcal{G}\}$ }
\For{sampled batch $\{g_k\}_{k=1}^{N} \in \mathcal{G}$}{
\footnotesize\tcp{compute encodings:}\normalsize
\For{$k=1$ to $N$}{
$\mathbf{X}_k, \mathbf{A}_k = \Gamma(g_k)$   \scriptsize\tcp*[r]{sub-sample graph}\normalsize

$\mathbf{V}_k^{\alpha} = \tau_\alpha\left(\mathbf{A}_k\right)$ \scriptsize\tcp*[r]{first view}\normalsize
$\mathbf{Z}^\alpha_k=g_\theta \left(\mathbf{X}_k, \mathbf{V}_k^\alpha\right)$ \scriptsize\tcp*[r]{node rep.}\normalsize
$\mathbf{H}^\alpha_k=f_\psi\left(\mathbf{Z}^\alpha_k\right)$ \scriptsize\tcp*[r]{projected node rep.}\normalsize
$\vec{h}_k^\alpha = f_\phi\left(\mathcal{P}\left(\mathbf{Z}^\alpha_k\right)\right)$\scriptsize\tcp*[r]{projected graph rep.}\normalsize

$\mathbf{V}_k^{\beta} = \tau_\beta\left(\mathbf{A}_k\right)$  \scriptsize\tcp*[r]{second view}\normalsize
$\mathbf{Z}^\beta_k=g_\omega \left(\mathbf{X}_k, \mathbf{V}_k^\beta\right)$ \scriptsize\tcp*[r]{node rep.}\normalsize
$\mathbf{H}^\beta_k=f_\psi\left(\mathbf{Z}^\beta_k\right)$ \scriptsize\tcp*[r]{projected node rep.}\normalsize
$\vec{h}_k^\beta = f_\phi\left(\mathcal{P}\left(\mathbf{Z}^\beta_k\right)\right)$ \scriptsize\tcp*[r]{projected graph rep.}\normalsize
}
\footnotesize\tcp{compute pairwise similarity:}\normalsize
\For{$i=1$ to $N$ and $j=1$ to $N$}{
$s_{ij}^{\alpha} = \mathcal{D}\left(\vec{h}_i^\alpha, \mathbf{H}^\beta_j\right)$, $s_{ij}^{\beta} = \mathcal{D}\left(\vec{h}_i^\beta, \mathbf{H}^\alpha_j\right)$
}
\footnotesize\tcp{compute gradients:}\normalsize
$\bigtriangledown_{\theta, \omega, \phi, \psi} \frac{1}{N^2} \sum\limits_{i=1}^{N} \sum\limits_{j=1}^N{ \left[\mathcal{L}\left(s_{ij}^{\alpha}\right) + \mathcal{L}\left(s_{ij}^{\beta}\right)\right]} $
}	
\KwRet{$\left[\mathbf{H}^\alpha_g+\mathbf{H}^\beta_g, \vec{h}_g^\alpha +\vec{h}_g^\beta \right]$,  $\forall  g \in \mathcal{G}$ }
\caption{Contrastive multi-view graph representation learning algorithm.} \label{algo}
\end{algorithm}
\section{Experimental Results}
\begin{table*}
\caption{Statistics of classification benchmarks.}
\label{table:datastat}
\setlength{\tabcolsep}{4pt}
\vskip 0.05in
\begin{center}
\begin{small}
\begin{sc}
\begin{tabular}{lccc|ccccc}
\toprule
& \multicolumn{3}{c}{\textit{Node}} & \multicolumn{5}{c}{\textit{Graph}}\\
\cline{2-9}
& \textbf{cora} & \textbf{citeseer} & \textbf{pubmed} & \textbf{mutag} & \textbf{ptc-mr} & \textbf{imdb-bin} &\textbf{ imdb-multi} & \textbf{reddit-bin}\\
\midrule
$|$ Graphs$|$  & 1 & 1 & 1 & 188 & 344 & 1000 & 1500 & 2000\\
$|$ Nodes$|$  & 3,327 & 2,708 & 19,717 & 17.93 & 14.29 &  19.77 & 13.0 & 508.52\\
$|$ Edges$|$  & 4,732 & 5,429 & 44,338 & 19.79 & 14.69 & 193.06 & 65.93 & 497.75\\
$|$ Classes$|$  & 6 & 7 & 3 & 2 & 2 & 2 & 3 & 2\\
\bottomrule
\end{tabular}
\end{sc}
\end{small}
\end{center}
\vskip -0.2in
\end{table*}
\begin{table*}
\caption{Mean node classification accuracy for supervised and unsupervised models. The input column highlights the data available to each model
during training (\textbf{X}: features, \textbf{A}: adjacency matrix, \textbf{S}: diffusion matrix, \textbf{Y}: labels).}
\label{table:node}
\vskip 0.05in
\begin{center}
\begin{small}
\begin{sc}
\begin{tabular}{llllccc}
\toprule
\multicolumn{2}{c}{\textbf{method}} & \textbf{input} & \textbf{cora} & \textbf{citeseer} & \textbf{pubmed} \\
\midrule
\multirow{11}{*}{\begin{turn}{90}supervised\end{turn}}  
& MLP \cite{velickovic_2018_iclr}  & $\textbf{X}$, $\textbf{Y}$ & 55.1 & 46.5 & 71.4\\
& ICA \cite{lu_2003_icml} & $\textbf{A}$, $\textbf{Y}$ &75.1 & 69.1 & 73.9 \\
& LP \cite{zhu_2003_icml} & $\textbf{A}$, $\textbf{Y}$ &  68.0  & 45.3 & 63.0 \\
& ManiReg \cite{Belkin_2006_jmlr} &  $\textbf{X}$, $\textbf{A}$, $\textbf{Y}$ & 59.5 & 60.1  & 70.7 \\
& SemiEmb \cite{weston_2012_nn} &  $\textbf{X}$, $\textbf{Y}$  & 59.0 &  59.6 & 71.7 \\
& Planetoid \cite{yang_2016_icml} & $\textbf{X}$, $\textbf{Y}$ & 75.7 & 64.7 & 77.2\\
& Chebyshev \cite{defferrard_2016_nips} &  $\textbf{X}$, $\textbf{A}$, $\textbf{Y}$ & 81.2 & 69.8 & 74.4\\
& GCN \cite{kipf_2017_iclr}   & $\textbf{X}$, $\textbf{A}$, $\textbf{Y}$ & 81.5   & 70.3  & 79.0\\
& MoNet \cite{Monti_2017_cvpr} & $\textbf{X}$, $\textbf{A}$, $\textbf{Y}$ & 81.7 $\pm$ 0.5  & $-$ & 78.8 $\pm$ 0.3 \\
& JKNet \cite{Xu_2018_icml} & $\textbf{X}$, $\textbf{A}$, $\textbf{Y}$ &  82.7 $\pm$ 0.4 &  \textbf{73.0 $\pm$ 0.5} &  77.9 $\pm$ 0.4  \\
& GAT \cite{velickovic_2018_iclr} & $\textbf{X}$, $\textbf{A}$, $\textbf{Y}$ & \textbf{83.0 $\pm$ 0.7} & 72.5 $\pm$ 0.7 & \textbf{79.0 $\pm$ 0.3} \\
\midrule
\multirow{7}{*}{\begin{turn}{90}unsupervised\end{turn}}
& Linear \cite{velickovic_2019_iclr}  & $\textbf{X}$ & 47.9 $\pm$ 0.4 &  49.3 $\pm$  0.2 &  69.1 $\pm$  0.3\\
& DeepWalk \cite{perozzi_2014_kdd} & $\textbf{X}$, $\textbf{A}$ & 70.7 $\pm$ 0.6 & 51.4 $\pm$ 0.5 & 74.3 $\pm$ 0.9 \\
& GAE \cite{kipf_2016_arxiv} & $\textbf{X}$, $\textbf{A}$ & 71.5 $\pm$ 0.4 & 65.8  $\pm$ 0.4 & 72.1 $\pm$ 0.5 \\
& VERSE \cite{tsitsulin_2018_www}  &  $\textbf{X}$, $\textbf{S}$, $\textbf{A}$ & 72.5 $\pm$ 0.3 & 55.5  $\pm$ 0.4  & $-$ \\
& DGI \cite{velickovic_2019_iclr}  & $\textbf{X}$, $\textbf{A}$ & 82.3 $\pm$ 0.6 & 71.8 $\pm$ 0.7  & 76.8 $\pm$ 0.6 \\
& DGI \cite{velickovic_2019_iclr}  & $\textbf{X}$, $\textbf{S}$ & 83.8 $\pm$ 0.5  & 72.0 $\pm$ 0.6 & 77.9 $\pm$ 0.3 \\
& Ours  & $\textbf{X}$, $\textbf{S}$, $\textbf{A}$ & \underline{\textbf{86.8 $\pm$ 0.5}} & \underline{\textbf{73.3 $\pm$ 0.5}} & \underline{\textbf{80.1 $\pm$ 0.7}} \\
\bottomrule
\end{tabular}
\end{sc}
\end{small}
\end{center}
\vskip -0.2in
\end{table*}
\begin{table*}
\caption{Performance on node clustering task reported in normalized MI (NMI) and adjusted rand index (ARI) measures.}
\label{table:cluster}
\vskip 0.05in
\begin{center}
\begin{small}
\begin{sc}
\begin{tabular}{llcccccc}
\toprule
\multicolumn{2}{c}{\textbf{method}} & \multicolumn{2}{c}{\textbf{cora}} & \multicolumn{2}{c}{\textbf{citeseer}} 
& \multicolumn{2}{c}{\textbf{pubmed}} \\
\cline{3-8}
& & NMI & ARI & NMI & ARI & NMI & ARI \\
\midrule
\multirow{6}{*}{\begin{turn}{90}unsupervised\end{turn}}
& VGAE \cite{kipf_2016_arxiv} & 0.3292 & 0.2547 & 0.2605 & 0.2056 & 0.3108 & 0.3018 \\
& MGAE \cite {wang_2017_cikm} & 0.5111 & 0.4447 & 0.4122 & 0.4137 & 0.2822 & 0.2483 \\
& ARGA \cite{pan_2018_ijcai}  & 0.4490 & 0.3520 & 0.3500 & 0.3410 & 0.2757 & 0.2910 \\
& ARVGA \cite{pan_2018_ijcai} & 0.4500 & 0.3740 & 0.2610 & 0.2450 & 0.1169 & 0.0777 \\
& GALA \cite{park_2019_iccv} & 0.5767 & 0.5315 & 0.4411 & 0.4460 & 0.3273 & 0.3214 \\
& Ours & \textbf{0.6291} & \textbf{0.5696} & \textbf{0.4696} & \textbf{0.4497} & \textbf{0.3609} & \textbf{0.3386} \\
\bottomrule
\end{tabular}
\end{sc}
\end{small}
\end{center}
\vskip -0.2in
\end{table*}
\begin{table*}
\caption{Mean 10-fold cross validation accuracy on graphs for kernel, supervised, and unsupervised methods.}
\label{table:graph}
\vskip 0.05in
\begin{center}
\begin{small}
\begin{sc}
\begin{tabular}{clcccccc}
\toprule
\multicolumn{2}{c}{\textbf{method}} & \textbf{mutag} & \textbf{ptc-mr} & \textbf{imdb-bin} &\textbf{ imdb-multi} & \textbf{reddit-bin} \\
\midrule
\multirow{5}{*}{\begin{turn}{90}Kernel\end{turn}}  
& SP \cite{borgwardt_2005_icdm}  & 85.2 $\pm$ 2.4 & 58.2 $\pm$ 2.4  & 55.6 $\pm$ 0.2 & 38.0 $\pm$ 0.3 & 64.1 $\pm$ 0.1\\
& GK \citep{shervashidze_2009_ais} & 81.7 $\pm$ 2.1 & 57.3 $\pm$ 1.4  & 65.9 $\pm$ 1.0 & 43.9 $\pm$ 0.4 & 77.3 $\pm$ 0.2\\
& WL \cite{shervashidze_2011_jmlr} & 80.7 $\pm$ 3.0 & 58.0 $\pm$ 0.5  & \textbf{72.3 $\pm$ 3.4} & \textbf{47.0 $\pm$ 0.5} & 68.8 $\pm$ 0.4\\
& DGK \cite{yanardag_2015_kdd} & 87.4 $\pm$ 2.7 & 60.1 $\pm$ 2.6  & 67.0 $\pm$ 0.6 & 44.6 $\pm$ 0.5 & \textbf{78.0 $\pm$ 0.4}\\
& MLG \cite{kondor_2016_nips} & \textbf{87.9 $\pm$ 1.6} & \textbf{63.3 $\pm$ 1.5} & 66.6 $\pm$ 0.3 & 41.2 $\pm$ 0.0 & $-$\\
\midrule
\multirow{5}{*}{\begin{turn}{90}supervised\end{turn}}
& GraphSAGE \cite{hamilton_2017_nips} & 85.1  $\pm$  7.6 & 63.9  $\pm$  7.7 & 72.3  $\pm$  5.3 & 50.9  $\pm$  2.2 & $–$  \\ 
& GCN \cite{kipf_2017_iclr} & 85.6  $\pm$  5.8 & 64.2  $\pm$  4.3 & 74.0  $\pm$  3.4 & 51.9  $\pm$  3.8 & 50.0  $\pm$  0.0 \\
& GIN-0 \cite{xu_2019_iclr} & \textbf{89.4  $\pm$  5.6} & 64.6  $\pm$  7.0 & \underline{\textbf{75.1  $\pm$  5.1}} &\underline{\textbf{52.3  $\pm$  2.8}} & \underline{\textbf{92.4  $\pm$  2.5}} \\
& GIN-$\epsilon$ \cite{xu_2019_iclr} & 89.0  $\pm$  6.0 & 63.7  $\pm$  8.2 & 74.3  $\pm$  5.1 & 52.1  $\pm$  3.6 & 92.2  $\pm$  2.3 \\
& GAT \cite{velickovic_2018_iclr} & \textbf{89.4 $\pm$ 6.1} & \underline{\textbf{66.7 $\pm$ 5.1}} & 70.5 $\pm$ 2.3 & 47.8 $\pm$ 3.1 & 85.2 $\pm$ 3.3 \\
\midrule
\multirow{6}{*}{\begin{turn}{90}unsupervised\end{turn}}
& random walk \cite{gartner_2003_ltkm} & 83.7 $\pm$ 1.5 & 57.9 $\pm$ 1.3 &  50.7 $\pm$ 0.3 & 34.7 $\pm$ 0.2 & $-$\\
& node2vec \cite{grover_2016_kdd}  & 72.6 $\pm$ 10.2  & 58.6 $\pm$ 8.0  & $-$  & $-$  & $-$ \\
& sub2vec \cite{adhikari_2018_pakdd} & 61.1 $\pm$ 15.8 & 60.0 $\pm$ 6.4 & 55.3 $\pm$ 1.5 & 36.7 $\pm$ 0.8 & 71.5 $\pm$ 0.4\\
& graph2vec \cite{narayanan_2017_arxiv} & 83.2 $\pm$ 9.6 & 60.2 $\pm$ 6.9 &  71.1 $\pm$ 0.5 & 50.4 $\pm$ 0.9  & 75.8 $\pm$ 1.0\\
& InfoGraph \cite{Sun_2020_iclr} & 89.0 $\pm$ 1.1 & 61.7 $\pm$ 1.4 & 73.0 $\pm$ 0.9 & 49.7 $\pm$ 0.5  & 82.5 $\pm$ 1.4\\
& Ours & \underline{\textbf{89.7 $\pm$ 1.1}}  &  \textbf{62.5 $\pm$ 1.7} & \textbf{74.2 $\pm$ 0.7} & \textbf{51.2 $\pm$ 0.5} & \textbf{84.5 $\pm$ 0.6}\\
\bottomrule
\end{tabular}
\end{sc}
\end{small}
\end{center}
\vskip -0.2in
\end{table*}
\subsection{Benchmarks}
We use three node classification and five graph classification benchmarks widely used in the literature \cite{kipf_2017_iclr, velickovic_2018_iclr,velickovic_2019_iclr, 
Sun_2020_iclr}. For node classification, we use Citeseer, Cora, and Pubmed citation networks \cite{sen_2008_aimag} where documents (nodes) are connected 
through citations (edges). For graph classification, we use the following: MUTAG \cite{kriege_2012_icml} containing mutagenic compounds, PTC 
\cite{kriege_2012_icml} containing compounds tested for carcinogenicity, Reddit-Binary \cite{yanardag_2015_kdd} connecting users (nodes) through responses 
(edges) in Reddit online discussions, and IMDB-Binary and IMDB-Multi \cite{yanardag_2015_kdd} connecting actors/actresses (nodes) based on movie 
appearances (edges). The statistics are summarized in Table \ref{table:datastat}.
\subsection{Evaluation Protocol}
For both node and graph classification benchmarks, we evaluate the proposed approach under the linear evaluation protocol and for each task, we closely follow
the experimental protocol of the previous state-of-the-art approaches. For node classification, we follow DGI and report the mean classification accuracy with 
standard deviation on the test nodes after 50 runs of training followed by a linear model. For graph classification, we follow InfoGraph and report the mean 
10-fold cross validation accuracy with standard deviation after 5 runs followed by a linear SVM. The linear classifier is trained using cross validation on training 
folds of data and the best mean classification accuracy is reported. Moreover, for node classification benchmarks, we evaluate the proposed method under 
clustering evaluation protocol and cluster the learned representations using K-Means algorithm. Similar to \cite{park_2019_iccv}, we set the number of clusters to 
the number of ground-truth classes and report the average normalized MI (NMI) and adjusted rand index (ARI) for 50 runs. 

We initialize the parameters using Xavier initialization \cite{glorot_2010_aistat} and train the model using Adam optimizer \cite{kingma_2014_iclr} with an initial 
learning rate of 0.001. To have fair comparisons, we follow InfoGraph for graph classification and choose the number of GCN layers, number of epochs, batch size, 
and the C parameter of the SVM from [2, 4, 8, 12], [10, 20, 40, 100], [32, 64, 128, 256], and $ [10^{-3}$, $10^{-2}$, ..., $10^2$, $10^3$], respectively. For node classification, we follow 
DGI and set the number of GCN layers and the number of epochs and to 1 and 2,000, respectively, and choose the batch size from [2, 4, 8]. We also use early 
stopping with a patience of 20. Finally, we set the size of hidden dimension of both node and graph representations to 512. For chosen hyper-parameters see Appendix.
\subsection{Comparison with State-of-the-Art}
To evaluate node classification under the linear evaluation protocol, we compare our results with unsupervised models including DeepWalk 
\cite{perozzi_2014_kdd} and DGI. We also train a GAE \cite{kipf_2016_arxiv}, a variant of DGI with a GDC encoder, and a variant of VERSE \cite{tsitsulin_2018_www} 
by minimizing KL-divergence between node representations and diffusion matrix. Furthermore, we compare our results with supervised models including an 
MLP, iterative classification algorithm (ICA) \cite{lu_2003_icml}, label propagation (LP) \cite{zhu_2003_icml}, manifold regularization (ManiReg) 
\cite{Belkin_2006_jmlr}, semi-supervised embedding (SemiEmb) \cite{weston_2012_nn}, Planetoid \cite{yang_2016_icml}, Chebyshev \cite{defferrard_2016_nips}, 
mixture model networks (MoNet) \cite{Monti_2017_cvpr}, JK-Net, GCN, and GAT. The results reported in Table \ref{table:node} show that we achieve 
state-of-the-art results with respect to previous unsupervised models. For example, on Cora, we achieve 86.8\% accuracy, which is a 5.5\% relative improvement 
over previous state-of-the-art. When compared to supervised baselines, we outperform strong supervised baselines: on Cora and PubMed benchmarks we observe 
4.5\% and 1.4\% relative improvement over GAT, respectively. 

To evaluate node classification under the clustering protocol, we compare our model with models reported in \cite{park_2019_iccv} including: variational GAE
(VGAE) \cite{kipf_2016_arxiv}, marginalized GAE (MGAE) \cite {wang_2017_cikm}, adversarially regularized GAE (ARGA) and VGAE (ARVGA) 
\cite{pan_2018_ijcai}, and GALA \cite{park_2019_iccv}. The results shown in Table \ref{table:cluster} suggest that our model achieves state-of-the-art NMI and ARI 
scores across all benchmarks.

To evaluate graph classification under the linear evaluation protocol, we compare our results with five graph kernel methods including shortest path kernel 
(SP) \cite{borgwardt_2005_icdm}, Graphlet kernel (GK) \cite{shervashidze_2009_ais}, Weisfeiler-Lehman sub-tree kernel (WL) \cite{shervashidze_2011_jmlr}, deep 
graph kernels (DGK) \cite{yanardag_2015_kdd}, and multi-scale Laplacian kernel (MLG) \cite{kondor_2016_nips} reported in \cite {Sun_2020_iclr}. We also compare 
with five supervised GNNs reported in \cite{xu_2019_iclr} including GraphSAGE \cite{hamilton_2017_nips}, GCN, GAT, and two variants of GIN: GIN-0 and 
GIN-$\epsilon$. Finally, We compare the results with five unsupervised methods including random walk \cite{gartner_2003_ltkm}, node2vec \cite{grover_2016_kdd}, sub2vec 
\cite{adhikari_2018_pakdd}, graph2vec \cite{narayanan_2017_arxiv}, and InfoGraph. The results shown in Table \ref{table:graph} suggest that our approach achieves
state-of-the-art results with respect to unsupervised models. For example, on Reddit-Binary \cite{yanardag_2015_kdd}, it achieves 84.5\% accuracy, i.e., a 2.4\% 
relative improvement over previous state-of-the-art. Our model also outperforms kernel methods in 4 out of 5 datasets and also outperforms best supervised 
model in one of the datasets. When compared to supervised baselines individually, our model outperforms GCN and GAT models in 3 out of 5 datasets, e.g., a 
5.3\% relative improvement over GAT on IMDB-Binary dataset. 

It is noteworthy that we achieve the state-of-the-art results on both node and graph classification benchmarks using a unified approach and unlike previous  unsupervised models \cite{velickovic_2019_iclr, Sun_2020_iclr}, we do not devise a specialized encoder for each task.
\subsection{Ablation Study }
\label{sec:ablation}
\begin{table*}
\caption{Effect of MI estimator, contrastive mode, and views on the accuracy on both node and graph classification tasks.}
\label{table:ablation}
\setlength{\tabcolsep}{3.5pt}
\vskip -0.1in
\begin{center}
\begin{small}
\begin{sc}
\begin{tabular}{clccc|ccccc}
\toprule
& ~ & \multicolumn{3}{c}{\textit{Node}} & \multicolumn{5}{c}{\textit{Graph}}\\
\cline{3-10}
& ~ & \textbf{cora} & \textbf{citeseer} & \textbf{pubmed} & \textbf{mutag} & \textbf{ptc-mr} & \textbf{imdb-bin} &\textbf{ imdb-multi} & \textbf{reddit-bin} \\
\midrule
\multirow{4}{*}{\begin{turn}{90}MI Est.\end{turn}}
& NCE         & 85.8 $\pm$ 0.7 & \textbf{73.3 $\pm$ 0.5} & \textbf{80.1 $\pm$ 0.7} & 82.2 $\pm$ 3.2 & 54.6 $\pm$ 2.5   &  73.7 $\pm$ 0.5   &  50.8 $\pm$ 0.8   &  79.7 $\pm$ 2.2 \\
& JSD           & 86.7 $\pm$ 0.6 & 72.9 $\pm$ 0.6 & 79.4 $\pm$ 1.0 &  \textbf{89.7 $\pm$ 1.1}  &  \textbf{62.5 $\pm$ 1.7}  &  \textbf{74.2 $\pm$ 0.7}   &  \textbf{51.1 $\pm$ 0.5}   &  \textbf{84.5 $\pm$ 0.6} \\
& NT-Xent   & \textbf{86.8 $\pm$ 0.5} & 72.9 $\pm$ 0.6 & 79.3 $\pm$ 0.8 & 75.4 $\pm$ 7.8   &   51.2 $\pm$ 3.3   &   63.6 $\pm$ 4.2   &   50.4 $\pm$ 0.6   &   82.0 $\pm$ 1.1 \\
& DV            & 85.4 $\pm$ 0.6 & \textbf{73.3 $\pm$ 0.5} & 78.9 $\pm$ 0.8 &  83.4 $\pm$ 1.9  &  56.7 $\pm$ 2.5  &  72.5 $\pm$ 0.8  & \textbf{51.1 $\pm$ 0.5}   &  76.3 $\pm$ 5.6 \\
\midrule
\multirow{5}{*}{\begin{turn}{90}Mode\end{turn}}
& local-global & \textbf{86.8 $\pm$ 0.5}  & \textbf{73.3 $\pm$ 0.5}  & \textbf{80.1 $\pm$ 0.7} &\textbf{ 89.7 $\pm$ 1.1} & \textbf{62.5 $\pm$ 1.7} & \textbf{74.2 $\pm$ 0.7} &  \textbf{51.1 $\pm$ 0.5} & \textbf{84.5 $\pm$ 0.6} \\ 
& global & $-$ & $-$ & $-$ &  85.4 $\pm$ 2.8 & 56.0 $\pm$ 2.1 & 72.4 $\pm$ 0.4 & 49.7 $\pm$ 0.8 & 80.8 $\pm$ 1.8 \\ 
& Multi-scale & 83.2 $\pm$ 0.9 & 63.5 $\pm$ 1.5 & 75.7 $\pm$ 1.1 & 88.0 $\pm$ 0.8 & 56.6 $\pm$ 1.8 & 72.7 $\pm$ 0.4 & 50.6 $\pm$ 0.5 & 82.8 $\pm$ 0.6 \\ 
& Hybrid  & $-$ & $-$ & $-$ &  86.1 $\pm$ 1.7  & 56.1 $\pm$ 1.4  & 73.3 $\pm$ 1.2  & 49.6 $\pm$ 0.6  & 78.2 $\pm$ 4.2  \\ 
& Ensemble  & 86.2 $\pm$ 0.6 & \textbf{73.3 $\pm$ 0.5} & 79.7 $\pm$ 0.9 &  82.5 $\pm$ 1.9 & 54.0 $\pm$ 3.0 & 73.0 $\pm$ 0.4 & 49.9 $\pm$ 0.9 & 81.4 $\pm$ 1.8 \\
\midrule
\multirow{6}{*}{\begin{turn}{90}Views\end{turn}}
& Adj-ppr & \textbf{86.8 $\pm$ 0.5}   & \textbf{73.3 $\pm$ 0.5}   &\textbf{ 80.1 $\pm$ 0.7} & \textbf{89.7 $\pm$ 1.1} & \textbf{62.5 $\pm$ 1.7} & \textbf{74.2 $\pm$ 0.7} & 51.1 $\pm$ 0.5 & \textbf{84.5 $\pm$ 0.6} \\
& Adj-heat & 86.4 $\pm$ 0.5  & 71.8 $\pm$ 0.5 & 77.2 $\pm$ 1.2 & 85.0 $\pm$ 1.9 & 55.8 $\pm$ 1.1 & 72.8 $\pm$ 0.5 & 50.0 $\pm$ 0.6 & 81.6 $\pm$ 0.9 \\
& Adj-dist & 84.5 $\pm$ 0.6 & 72.7 $\pm$ 0.7 & 74.6 $\pm$ 1.4 & 87.1 $\pm$ 1.0 & 58.7 $\pm$ 2.2 & 72.0 $\pm$ 0.7 & 50.7 $\pm$ 0.6 & 81.8 $\pm$ 0.7 \\
& ppr-heat & 85.8 $\pm$ 0.5 & 72.9 $\pm$ 0.5 & 78.1 $\pm$ 0.9 & 87.7 $\pm$ 1.2 & 57.6 $\pm$ 1.6 & 72.2 $\pm$ 0.6 & \textbf{51.2 $\pm$ 0.8} & 82.3 $\pm$ 1.0 \\
& ppr-dist & 85.9 $\pm$ 0.5 & 73.2 $\pm$ 0.4 & 74.7 $\pm$ 1.2 & 87.1 $\pm$ 1.0 & 60.0 $\pm$ 2.5 & 72.4 $\pm$ 1.4 & 51.1 $\pm$ 0.8 & 82.5 $\pm$ 1.1 \\
& heat-dist & 85.2 $\pm$ 0.4 & 70.4 $\pm$ 0.7 & 72.8 $\pm$ 0.7 & 87.4 $\pm$ 1.2 & 58.6 $\pm$ 1.7 & 72.2 $\pm$ 0.6 & 50.5 $\pm$ 0.5 & 80.3 $\pm$ 0.6 \\
\bottomrule
\end{tabular}
\end{sc}
\end{small}
\end{center}
\vskip -0.2in
\end{table*}
\subsubsection{Effect of Mutual Information Estimator}
We investigated four MI estimators including: noise-contrastive estimation (NCE) \cite{gutmann_2010_aistat, oord_2018_arxiv}, Jensen-Shannon (JSD) estimator 
following formulation in \cite{nowozin_2016_nips}, normalized temperature-scaled cross-entropy (NT-Xent) \cite{chen_2020_arxiv}, and Donsker-Varadhan (DV) 
representation of the KL-divergence \cite{donsker_1975_comm}. The results shown in Table \ref{table:ablation} suggests that Jensen-Shannon estimator achieves
better results across all graph classification benchmarks, whereas in node classification benchmarks, NCE achieves better results in 2 out of 3 datasets.

\subsubsection{Effect of Contrastive Mode }
We considered five contrastive modes including: local-global, global-global, multi-scale, hybrid, and ensemble modes. In local-global mode we extend deep 
InfoMax \cite{hjelm_2019_iclr} and contrast node encodings from one view with graph encodings from another view and vice versa. Global-global mode is similar
to \cite{li_2019_icml, tian_2019_arxiv, chen_2020_arxiv} where we contrast graph encodings from different views. In multi-scale mode, we contrast graph encodings 
from one view with intermediate encodings from another view and vice versa, and we also contrast intermediate encodings from one view with node encodings 
from another view and vice versa. We use two DiffPool layers to compute the intermediate encodings. The first layer projects nodes into a set of clusters where the 
number of clusters, i.e., motifs, is set as 25\% of the number of nodes before applying DiffPool, whereas the second layer projects the learned cluster encodings 
into a graph encoding. In hybrid mode, we use both local-global and global-global modes. Finally, in ensemble mode, we contrast nodes and graph encodings from a same view for all views.

Results reported in Table \ref{table:ablation} suggest that contrasting node and graph encodings consistently perform better across benchmarks. The results also 
reveal important differences between graph and visual representation learning: (1) in visual representation learning, contrasting global views achieves best results
\cite{tian_2019_arxiv, chen_2020_arxiv}, whereas for graphs, contrasting node and graph encodings achieves better performance for both node and graph 
classification tasks, and (2) contrasting multi-scale encodings helps visual representation learning \cite{bachman_2019_nips} but has a negative effect on graph representation learning.

\subsubsection{Effect of Views}
We investigated four structural views including adjacency matrix, PPR and heat diffusion matrices, and pair-wise distance matrix where adjacency conveys local 
information and the latter three capture global information. Adjacency matrix is processed into a symmetrically normalized adjacency matrix (see section 
\ref{encoder}), and PPR and heat diffusion matrices are computed using Eq. (\ref{eq:ppr}) and (\ref{eq:heat}), respectively. The shortest pair-wise distance matrix is 
computed by Floyd-Warshall algorithm, inversed in an element-wise fashion, and row-normalized using softmax. All views are computed once in preprocessing. 
The results shown in Table \ref{table:ablation} suggest that contrasting encodings from adjacency and PPR views performs better across the benchmarks.

Furthermore, we investigated whether increasing the number of views increases the performance on down-stream tasks, monotonically. Following 
\cite{tian_2019_arxiv}, we extended number of views to three by anchoring the main view on adjacency matrix and considering two diffusion matrices as other 
views. The results (see Appendix) suggest that unlike visual representation learning, extending the views does not help. We speculate this is because different 
diffusion matrices carry similar information about the graph structure. We also followed \cite{chen_2020_arxiv} and randomly sampled 2 out of 4 views for each 
training sample in each mini-batch and contrasted the representation. We observed that unlike visual domain, it degraded the performance.

\subsubsection{Negative Sampling \& Regularization}
We investigated the performance with respect to batch size where a batch of size $N$ consists of $N-1$  negative and 1 positive examples. We observed that in graph classification, increasing the batch size slightly improves the performance, whereas, in node classification, it does not have a significant effect. Thus we opted for 
efficient smaller batch sizes. To generate negative samples in node classification, we considered two corruption functions: (1) random feature permutation, and (2)
adjacency matrix corruption. We observed that applying the former achieves significantly better results compared to the latter or a combination of both.

Furthermore, we observed that applying normalization layers such as BatchNorm \cite{ioffe_2015_icml} or LayerNorm \cite{ba_2016_arxiv}, or
regularization methods such as adding Gaussian noise, L2 regularization, or dropout \cite{srivastava_2014_jmlr} during the pre-training degrades the performance on down-stream tasks (except early stopping).

\section{Conclusion}
We introduced a self-supervised approach for learning node and graph level representations by contrasting encodings from two structural views of graphs including first-order neighbors and a graph diffusion. We showed that unlike visual representation learning, increasing the number of views or contrasting multi-scale encodings do not improve the performance. Using these findings, we achieved new state-of-the-art in self-supervised learning on 8 out of 8 node and graph 
classification benchmarks under the linear evaluation protocol and outperformed strong supervised baselines in 4 out of 8 benchmarks. In future work, we are
planning to investigate large pre-training and transfer learning capabilities of the proposed method.

\small
\bibliography{paper}

\begin{thebibliography}{71}
\providecommand{\natexlab}[1]{#1}
\providecommand{\url}[1]{\texttt{#1}}
\expandafter\ifx\csname urlstyle\endcsname\relax
  \providecommand{\doi}[1]{doi: #1}\else
  \providecommand{\doi}{doi: \begingroup \urlstyle{rm}\Url}\fi

\bibitem[Adhikari et~al.(2018)Adhikari, Zhang, Ramakrishnan, and
  Prakash]{adhikari_2018_pakdd}
Adhikari, B., Zhang, Y., Ramakrishnan, N., and Prakash, B.~A.
\newblock Sub2vec: Feature learning for subgraphs.
\newblock In \emph{Pacific-Asia Conference on Knowledge Discovery and Data
  Mining}, pp.\  170--182, 2018.

\bibitem[Ba et~al.(2016)Ba, Kiros, and Hinton]{ba_2016_arxiv}
Ba, J.~L., Kiros, J.~R., and Hinton, G.~E.
\newblock Layer normalization.
\newblock \emph{arXiv preprint arXiv:1607.06450}, 2016.

\bibitem[Bachman et~al.(2019)Bachman, Hjelm, and Buchwalter]{bachman_2019_nips}
Bachman, P., Hjelm, R.~D., and Buchwalter, W.
\newblock Learning representations by maximizing mutual information across
  views.
\newblock In \emph{Advances in Neural Information Processing Systems}, pp.\
  15509--15519, 2019.

\bibitem[Belkin et~al.(2006)Belkin, Niyogi, and Sindhwani]{Belkin_2006_jmlr}
Belkin, M., Niyogi, P., and Sindhwani, V.
\newblock Manifold regularization: A geometric framework for learning from
  labeled and unlabeled examples.
\newblock \emph{Journal of Machine Learning Research}, 7:\penalty0 2399–2434,
  2006.

\bibitem[Borgwardt \& Kriegel(2005)Borgwardt and Kriegel]{borgwardt_2005_icdm}
Borgwardt, K.~M. and Kriegel, H.-P.
\newblock Shortest-path kernels on graphs.
\newblock In \emph{International Conference on Data Mining}, 2005.

\bibitem[Chen et~al.(2020)Chen, Kornblith, Norouzi, and
  Hinton]{chen_2020_arxiv}
Chen, T., Kornblith, S., Norouzi, M., and Hinton, G.
\newblock A simple framework for contrastive learning of visual
  representations.
\newblock \emph{arXiv preprint arXiv:2002.05709}, 2020.

\bibitem[Defferrard et~al.(2016)Defferrard, Bresson, and
  Vandergheynst]{defferrard_2016_nips}
Defferrard, M., Bresson, X., and Vandergheynst, P.
\newblock Convolutional neural networks on graphs with fast localized spectral
  filtering.
\newblock In \emph{Advances in Neural Information Processing Systems}, pp.\
  3844--3852. 2016.

\bibitem[Donsker \& Varadhan(1975)Donsker and Varadhan]{donsker_1975_comm}
Donsker, M.~D. and Varadhan, S.~S.
\newblock Asymptotic evaluation of certain markov process expectations for
  large time.
\newblock \emph{Communications on Pure and Applied Mathematics}, 28\penalty0
  (1):\penalty0 1--47, 1975.

\bibitem[Duvenaud et~al.(2015)Duvenaud, Maclaurin, Iparraguirre, Bombarell,
  Hirzel, Aspuru-Guzik, and Adams]{duvenaud_2015_nips}
Duvenaud, D.~K., Maclaurin, D., Iparraguirre, J., Bombarell, R., Hirzel, T.,
  Aspuru-Guzik, A., and Adams, R.~P.
\newblock Convolutional networks on graphs for learning molecular fingerprints.
\newblock In \emph{Advances in Neural Information Processing Systems}, pp.\
  2224--2232, 2015.

\bibitem[Garcia~Duran \& Niepert(2017)Garcia~Duran and
  Niepert]{duran_2017_nips}
Garcia~Duran, A. and Niepert, M.
\newblock Learning graph representations with embedding propagation.
\newblock In \emph{Advances in Neural Information Processing Systems}, pp.\
  5119--5130. 2017.

\bibitem[G{\"a}rtner et~al.(2003)G{\"a}rtner, Flach, and
  Wrobel]{gartner_2003_ltkm}
G{\"a}rtner, T., Flach, P., and Wrobel, S.
\newblock On graph kernels: Hardness results and efficient alternatives.
\newblock In \emph{Learning theory and kernel machines}, pp.\  129--143. 2003.

\bibitem[Gilmer et~al.(2017)Gilmer, Schoenholz, Riley, Vinyals, and
  Dahl]{gilmer_2017_icml}
Gilmer, J., Schoenholz, S.~S., Riley, P.~F., Vinyals, O., and Dahl, G.~E.
\newblock Neural message passing for quantum chemistry.
\newblock In \emph{International Conference on Machine Learning}, pp.\
  1263--1272, 2017.

\bibitem[Glorot \& Bengio(2010)Glorot and Bengio]{glorot_2010_aistat}
Glorot, X. and Bengio, Y.
\newblock Understanding the difficulty of training deep feedforward neural
  networks.
\newblock In \emph{International Conference on Artificial Intelligence and
  Statistics}, pp.\  249--256, 2010.

\bibitem[Grover \& Leskovec(2016)Grover and Leskovec]{grover_2016_kdd}
Grover, A. and Leskovec, J.
\newblock node2vec: Scalable feature learning for networks.
\newblock In \emph{International Conference on Knowledge Discovery and Data
  Mining}, pp.\  855--864, 2016.

\bibitem[Gutmann \& Hyvärinen(2010)Gutmann and
  Hyvärinen]{gutmann_2010_aistat}
Gutmann, M. and Hyvärinen, A.
\newblock Noise-contrastive estimation: A new estimation principle for
  unnormalized statistical models.
\newblock In \emph{International Conference on Artificial Intelligence and
  Statistics}, pp.\  297--304, 2010.

\bibitem[Hamilton et~al.(2017)Hamilton, Ying, and Leskovec]{hamilton_2017_nips}
Hamilton, W., Ying, Z., and Leskovec, J.
\newblock Inductive representation learning on large graphs.
\newblock In \emph{Advances in Neural Information Processing Systems}, pp.\
  1024--1034, 2017.

\bibitem[Hassani \& Haley(2019)Hassani and Haley]{hassani_2019_iccv}
Hassani, K. and Haley, M.
\newblock Unsupervised multi-task feature learning on point clouds.
\newblock In \emph{International Conference on Computer Vision}, pp.\
  8160--8171, 2019.

\bibitem[He et~al.(2015)He, Zhang, Ren, and Sun]{he_2015_iccv}
He, K., Zhang, X., Ren, S., and Sun, J.
\newblock Delving deep into rectifiers: Surpassing human-level performance on
  imagenet classification.
\newblock In \emph{International Conference on Computer Vision}, pp.\
  1026--1034, 2015.

\bibitem[Hjelm et~al.(2019)Hjelm, Fedorov, Lavoie-Marchildon, Grewal, Bachman,
  Trischler, and Bengio]{hjelm_2019_iclr}
Hjelm, R.~D., Fedorov, A., Lavoie-Marchildon, S., Grewal, K., Bachman, P.,
  Trischler, A., and Bengio, Y.
\newblock Learning deep representations by mutual information estimation and
  maximization.
\newblock In \emph{International Conference on Learning Representations}, 2019.

\bibitem[Ioffe \& Szegedy(2015)Ioffe and Szegedy]{ioffe_2015_icml}
Ioffe, S. and Szegedy, C.
\newblock Batch normalization: Accelerating deep network training by reducing
  internal covariate shift.
\newblock In \emph{International Conference on Machine Learning}, pp.\
  448–456, 2015.

\bibitem[Jiang et~al.(2019)Jiang, Lin, Tang, and Luo]{jiang_2019_cvpr}
Jiang, B., Lin, D., Tang, J., and Luo, B.
\newblock Data representation and learning with graph diffusion-embedding
  networks.
\newblock In \emph{Conference on Computer Vision and Pattern Recognition}, pp.\
   10414--10423, 2019.

\bibitem[Khasahmadi et~al.(2020)Khasahmadi, Hassani, Moradi, Lee, and
  Morris]{Khasahmadi_2020_iclr}
Khasahmadi, A., Hassani, K., Moradi, P., Lee, L., and Morris, Q.
\newblock Memory-based graph networks.
\newblock In \emph{International Conference on Learning Representations}, 2020.

\bibitem[Kingma \& Ba(2014)Kingma and Ba]{kingma_2014_iclr}
Kingma, D.~P. and Ba, J.~L.
\newblock Adam: Amethod for stochastic optimization.
\newblock In \emph{International Conference on Learning Representation}, 2014.

\bibitem[Kipf \& Welling(2016)Kipf and Welling]{kipf_2016_arxiv}
Kipf, T.~N. and Welling, M.
\newblock Variational graph auto-encoders.
\newblock \emph{arXiv preprint arXiv:1611.07308}, 2016.

\bibitem[Kipf \& Welling(2017)Kipf and Welling]{kipf_2017_iclr}
Kipf, T.~N. and Welling, M.
\newblock Semi-supervised classification with graph convolutional networks.
\newblock In \emph{International Conference on Learning Representations}, 2017.

\bibitem[Klicpera et~al.(2019{\natexlab{a}})Klicpera, Bojchevski, and
  Günnemann]{klicpera_2019_iclr}
Klicpera, J., Bojchevski, A., and Günnemann, S.
\newblock Combining neural networks with personalized pagerank for
  classification on graphs.
\newblock In \emph{International Conference on Learning Representations},
  2019{\natexlab{a}}.

\bibitem[Klicpera et~al.(2019{\natexlab{b}})Klicpera, Wei\ss~enberger, and
  G\"{u}nnemann]{klicpera_2019_nips}
Klicpera, J., Wei\ss~enberger, S., and G\"{u}nnemann, S.
\newblock Diffusion improves graph learning.
\newblock In \emph{Advances in Neural Information Processing Systems}, pp.\
  13333--13345. 2019{\natexlab{b}}.

\bibitem[Kondor \& Pan(2016)Kondor and Pan]{kondor_2016_nips}
Kondor, R. and Pan, H.
\newblock The multiscale laplacian graph kernel.
\newblock In \emph{Advances in Neural Information Processing Systems}, pp.\
  2990--2998. 2016.

\bibitem[Kondor \& Lafferty(2002)Kondor and Lafferty]{kondor_2002_icml}
Kondor, R.~I. and Lafferty, J.
\newblock Diffusion kernels on graphs and other discrete structures.
\newblock In \emph{International Conference on Machine Learning}, pp.\
  315--22, 2002.

\bibitem[Kriege \& Mutzel(2012)Kriege and Mutzel]{kriege_2012_icml}
Kriege, N. and Mutzel, P.
\newblock Subgraph matching kernels for attributed graphs.
\newblock In \emph{International Conference on Machine Learning}, pp.\
  291--298, 2012.

\bibitem[Kriege et~al.(2016)Kriege, Giscard, and Wilson]{kriege_2016_nips}
Kriege, N.~M., Giscard, P.-L., and Wilson, R.
\newblock On valid optimal assignment kernels and applications to graph
  classification.
\newblock In \emph{Advances in Neural Information Processing Systems}, pp.\
  1623--1631, 2016.

\bibitem[Li et~al.(2015)Li, Tarlow, Brockschmidt, and Zemel]{li_2015_iclr}
Li, Y., Tarlow, D., Brockschmidt, M., and Zemel, R.
\newblock Gated graph sequence neural networks.
\newblock In \emph{International Conference on Learning Representations}, 2015.

\bibitem[Li et~al.(2019)Li, Gu, Dullien, Vinyals, and Kohli]{li_2019_icml}
Li, Y., Gu, C., Dullien, T., Vinyals, O., and Kohli, P.
\newblock Graph matching networks for learning the similarity of graph
  structured objects.
\newblock In \emph{International Conference on Machine Learning}, pp.\
  3835--3845, 2019.

\bibitem[{Linsker}(1988)]{linsker_1988_computer}
{Linsker}, R.
\newblock Self-organization in a perceptual network.
\newblock \emph{Computer}, 21\penalty0 (3):\penalty0 105--117, 1988.

\bibitem[Lu \& Getoor(2003)Lu and Getoor]{lu_2003_icml}
Lu, Q. and Getoor, L.
\newblock Link-based classification.
\newblock In \emph{International Conference on Machine Learning}, pp.\
  496--503, 2003.

\bibitem[Monti et~al.(2017)Monti, Boscaini, Masci, Rodola, Svoboda, and
  Bronstein]{Monti_2017_cvpr}
Monti, F., Boscaini, D., Masci, J., Rodola, E., Svoboda, J., and Bronstein,
  M.~M.
\newblock Geometric deep learning on graphs and manifolds using mixture model
  cnns.
\newblock In \emph{Conference on Computer Vision and Pattern Recognition},
  2017.

\bibitem[Narayanan et~al.(2017)Narayanan, Chandramohan, Venkatesan, Chen, Liu,
  and Jaiswal]{narayanan_2017_arxiv}
Narayanan, A., Chandramohan, M., Venkatesan, R., Chen, L., Liu, Y., and
  Jaiswal, S.
\newblock graph2vec: Learning distributed representations of graphs.
\newblock \emph{arXiv preprint arXiv:1707.05005}, 2017.

\bibitem[Nowozin et~al.(2016)Nowozin, Cseke, and Tomioka]{nowozin_2016_nips}
Nowozin, S., Cseke, B., and Tomioka, R.
\newblock f-gan: Training generative neural samplers using variational
  divergence minimization.
\newblock In \emph{Advances in Neural Information Processing Systems}, pp.\
  271--279. 2016.

\bibitem[Oord et~al.(2018)Oord, Li, and Vinyals]{oord_2018_arxiv}
Oord, A. v.~d., Li, Y., and Vinyals, O.
\newblock Representation learning with contrastive predictive coding.
\newblock \emph{arXiv preprint arXiv:1807.03748}, 2018.

\bibitem[Page et~al.(1999)Page, Brin, Motwani, and
  Winograd]{page_1999_stanford}
Page, L., Brin, S., Motwani, R., and Winograd, T.
\newblock The pagerank citation ranking: Bringing order to the web.
\newblock Technical report, Stanford InfoLab, 1999.

\bibitem[Pan et~al.(2018)Pan, Hu, Long, Jiang, Yao, and Zhang]{pan_2018_ijcai}
Pan, S., Hu, R., Long, G., Jiang, J., Yao, L., and Zhang, C.
\newblock Adversarially regularized graph autoencoder for graph embedding.
\newblock In \emph{International Joint Conference on Artificial Intelligence},
  pp.\  2609--2615, 2018.

\bibitem[Park et~al.(2019)Park, Lee, Chang, Lee, and Choi]{park_2019_iccv}
Park, J., Lee, M., Chang, H.~J., Lee, K., and Choi, J.~Y.
\newblock Symmetric graph convolutional autoencoder for unsupervised graph
  representation learning.
\newblock In \emph{International Conference on Computer Vision}, pp.\
  6519--6528, 2019.

\bibitem[Perozzi et~al.(2014)Perozzi, Al-Rfou, and Skiena]{perozzi_2014_kdd}
Perozzi, B., Al-Rfou, R., and Skiena, S.
\newblock Deepwalk: Online learning of social representations.
\newblock In \emph{International Conference on Knowledge Discovery and Data
  Mining}, pp.\  701--710, 2014.

\bibitem[Ribeiro et~al.(2017)Ribeiro, Saverese, and
  Figueiredo]{ribeiro_2017_kdd}
Ribeiro, L., Saverese, P., and Figueiredo, D.
\newblock Struc2vec: Learning node representations from structural identity.
\newblock In \emph{International Conference on Knowledge Discovery and Data
  Mining}, pp.\  385–394, 2017.

\bibitem[Sanchez-Gonzalez et~al.(2018)Sanchez-Gonzalez, Heess, Springenberg,
  Merel, Riedmiller, Hadsell, and Battaglia]{sanchez_2018_icml}
Sanchez-Gonzalez, A., Heess, N., Springenberg, J.~T., Merel, J., Riedmiller,
  M., Hadsell, R., and Battaglia, P.
\newblock Graph networks as learnable physics engines for inference and
  control.
\newblock In \emph{International Conference on Machine Learning}, pp.\
  4470--4479, 2018.

\bibitem[Sen et~al.(2008)Sen, Namata, Bilgic, Getoor, Galligher, and
  Eliassi-Rad]{sen_2008_aimag}
Sen, P., Namata, G., Bilgic, M., Getoor, L., Galligher, B., and Eliassi-Rad, T.
\newblock Collective classification in network data.
\newblock \emph{AI Magazine}, 29\penalty0 (3):\penalty0 93--93, 2008.

\bibitem[Shervashidze et~al.(2009)Shervashidze, Vishwanathan, Petri, Mehlhorn,
  and Borgwardt]{shervashidze_2009_ais}
Shervashidze, N., Vishwanathan, S., Petri, T., Mehlhorn, K., and Borgwardt, K.
\newblock Efficient graphlet kernels for large graph comparison.
\newblock In \emph{Artificial Intelligence and Statistics}, pp.\  488--495,
  2009.

\bibitem[Shervashidze et~al.(2011)Shervashidze, Schweitzer, Leeuwen, Mehlhorn,
  and Borgwardt]{shervashidze_2011_jmlr}
Shervashidze, N., Schweitzer, P., Leeuwen, E. J.~v., Mehlhorn, K., and
  Borgwardt, K.~M.
\newblock Weisfeiler-lehman graph kernels.
\newblock \emph{Journal of Machine Learning Research}, 12:\penalty0 2539--2561,
  2011.

\bibitem[Srivastava et~al.(2014)Srivastava, Hinton, Krizhevsky, Sutskever, and
  Salakhutdinov]{srivastava_2014_jmlr}
Srivastava, N., Hinton, G., Krizhevsky, A., Sutskever, I., and Salakhutdinov,
  R.
\newblock Dropout: A simple way to prevent neural networks from overfitting.
\newblock \emph{Journal of Machine Learning Research}, 15\penalty0
  (56):\penalty0 1929--1958, 2014.

\bibitem[Sun et~al.(2020)Sun, Hoffman, Verma, and Tang]{Sun_2020_iclr}
Sun, F.-Y., Hoffman, J., Verma, V., and Tang, J.
\newblock Infograph: Unsupervised and semi-supervised graph-level
  representation learning via mutual information maximization.
\newblock In \emph{International Conference on Learning Representations}, 2020.

\bibitem[Tang et~al.(2015)Tang, Qu, Wang, Zhang, Yan, and Mei]{tang_2015_www}
Tang, J., Qu, M., Wang, M., Zhang, M., Yan, J., and Mei, Q.
\newblock Line: Large-scale information network embedding.
\newblock In \emph{International Conference on World Wide Web}, pp.\
  1067--1077, 2015.

\bibitem[Tian et~al.(2019)Tian, Krishnan, and Isola]{tian_2019_arxiv}
Tian, Y., Krishnan, D., and Isola, P.
\newblock Contrastive multiview coding.
\newblock \emph{arXiv preprint arXiv:1906.05849}, 2019.

\bibitem[Tschannen et~al.(2020)Tschannen, Djolonga, Rubenstein, Gelly, and
  Lucic]{tschannen_2020_iclr}
Tschannen, M., Djolonga, J., Rubenstein, P.~K., Gelly, S., and Lucic, M.
\newblock On mutual information maximization for representation learning.
\newblock In \emph{International Conference on Learning Representations}, 2020.

\bibitem[Tsitsulin et~al.(2018)Tsitsulin, Mottin, Karras, and
  M\"{u}ller]{tsitsulin_2018_www}
Tsitsulin, A., Mottin, D., Karras, P., and M\"{u}ller, E.
\newblock Verse: Versatile graph embeddings from similarity measures.
\newblock In \emph{International Conference on World Wide Web}, pp.\
  539–548, 2018.

\bibitem[Veličković et~al.(2018)Veličković, Cucurull, Casanova, Romero,
  Liò, and Bengio]{velickovic_2018_iclr}
Veličković, P., Cucurull, G., Casanova, A., Romero, A., Liò, P., and Bengio,
  Y.
\newblock Graph attention networks.
\newblock In \emph{International Conference on Learning Representations}, 2018.

\bibitem[Veličković et~al.(2019)Veličković, Fedus, Hamilton, Liò, Bengio,
  and Hjelm]{velickovic_2019_iclr}
Veličković, P., Fedus, W., Hamilton, W.~L., Liò, P., Bengio, Y., and Hjelm,
  R.~D.
\newblock Deep graph infomax.
\newblock In \emph{International Conference on Learning Representations}, 2019.

\bibitem[Vivona \& Hassani(2019)Vivona and Hassani]{vivona_2019_nips}
Vivona, S. and Hassani, K.
\newblock Relational graph representation learning for open-domain question
  answering.
\newblock \emph{Advances in Neural Information Processing Systems, Graph
  Representation Learning Workshop}, 2019.

\bibitem[Wang et~al.(2017)Wang, Pan, Long, Zhu, and Jiang]{wang_2017_cikm}
Wang, C., Pan, S., Long, G., Zhu, X., and Jiang, J.
\newblock Mgae: Marginalized graph autoencoder for graph clustering.
\newblock In \emph{Conference on Information and Knowledge Management}, pp.\
  889--898, 2017.

\bibitem[Wang et~al.(2018)Wang, Zhang, Li, Fu, Liu, and Jiang]{wang_2018_eccv}
Wang, N., Zhang, Y., Li, Z., Fu, Y., Liu, W., and Jiang, Y.-G.
\newblock Pixel2mesh: Generating 3d mesh models from single rgb images.
\newblock In \emph{European Conference on Computer Vision}, pp.\  52--67, 2018.

\bibitem[Wang et~al.(2019)Wang, Zhou, Fidler, and Ba]{wang_2018_iclr}
Wang, T., Zhou, Y., Fidler, S., and Ba, J.
\newblock Neural graph evolution: Automatic robot design.
\newblock In \emph{International Conference on Learning Representations}, 2019.

\bibitem[Weston et~al.(2012)Weston, Ratle, Mobahi, and
  Collobert]{weston_2012_nn}
Weston, J., Ratle, F., Mobahi, H., and Collobert, R.
\newblock Deep learning via semi-supervised embedding.
\newblock In \emph{Neural Networks: Tricks of the Trade}, pp.\  639--655. 2012.

\bibitem[Wu et~al.(2020)Wu, Pan, Chen, Long, Zhang, and Philip]{wu_2020_nnls}
Wu, Z., Pan, S., Chen, F., Long, G., Zhang, C., and Philip, S.~Y.
\newblock A comprehensive survey on graph neural networks.
\newblock \emph{IEEE Transactions on Neural Networks and Learning Systems},
  2020.

\bibitem[Xu et~al.(2019{\natexlab{a}})Xu, Shen, Cao, Cen, and
  Cheng]{xu_2019_ijcai}
Xu, B., Shen, H., Cao, Q., Cen, K., and Cheng, X.
\newblock Graph convolutional networks using heat kernel for semi-supervised
  learning.
\newblock In \emph{International Joint Conference on Artificial Intelligence},
  pp.\  1928--1934, 2019{\natexlab{a}}.

\bibitem[Xu et~al.(2018)Xu, Li, Tian, Sonobe, Kawarabayashi, and
  Jegelka]{Xu_2018_icml}
Xu, K., Li, C., Tian, Y., Sonobe, T., Kawarabayashi, K.-i., and Jegelka, S.
\newblock Representation learning on graphs with jumping knowledge networks.
\newblock In \emph{International Conference on Machine Learning}, pp.\
  5453--5462, 2018.

\bibitem[Xu et~al.(2019{\natexlab{b}})Xu, Hu, Leskovec, and
  Jegelka]{xu_2019_iclr}
Xu, K., Hu, W., Leskovec, J., and Jegelka, S.
\newblock How powerful are graph neural networks?
\newblock In \emph{International Conference on Learning Representations},
  2019{\natexlab{b}}.

\bibitem[Yanardag \& Vishwana(2015)Yanardag and Vishwana]{yanardag_2015_kdd}
Yanardag, P. and Vishwana, S.
\newblock Deep graph kernels.
\newblock In \emph{International Conference on Knowledge Discovery and Data
  Mining}, pp.\  1365--1374, 2015.

\bibitem[Yang et~al.(2016)Yang, Cohen, and Salakhudinov]{yang_2016_icml}
Yang, Z., Cohen, W., and Salakhudinov, R.
\newblock Revisiting semi-supervised learning with graph embeddings.
\newblock In \emph{International Conference on Machine Learning}, pp.\  40--48,
  2016.

\bibitem[Ying et~al.(2018)Ying, You, Morris, Ren, Hamilton, and
  Leskovec]{ying_2018_nips}
Ying, Z., You, J., Morris, C., Ren, X., Hamilton, W., and Leskovec, J.
\newblock Hierarchical graph representation learning with differentiable
  pooling.
\newblock In \emph{Advances in Neural Information Processing Systems}, pp.\
  4800--4810, 2018.

\bibitem[You et~al.(2019)You, Ying, and Leskovec]{you_2019_icml}
You, J., Ying, R., and Leskovec, J.
\newblock Position-aware graph neural networks.
\newblock In \emph{International Conference on Machine Learning}, pp.\
  7134--7143, 2019.

\bibitem[Zhang et~al.(2020)Zhang, Cui, and Zhu]{zhang_2020_kde}
Zhang, Z., Cui, P., and Zhu, W.
\newblock Deep learning on graphs: A survey.
\newblock \emph{IEEE Transactions on Knowledge and Data Engineering}, 2020.

\bibitem[Zhu et~al.(2003)Zhu, Ghahramani, and Lafferty]{zhu_2003_icml}
Zhu, X., Ghahramani, Z., and Lafferty, J.~D.
\newblock Semi-supervised learning using gaussian fields and harmonic
  functions.
\newblock In \emph{International Conference on Machine Learning}, pp.\
  912--919, 2003.

\end{thebibliography}
\bibliographystyle{icml2020}
\normalsize

\clearpage \onecolumn
\section{Appendix}
\subsection{Implementation Details}
The model is implemented with PyTorch and each benchmark is run on a single GPU. For a given dataset, the node features are processed as follows. If the dataset
contains initial node features, they are normalize using the standard score. If the dataset does not contains initial node features, but carries node labels, they are 
used as the initial node features (only in graph classification benchmarks). Otherwise, if the dataset has neither, the node features are initialized with node degrees.
Diffusion and pair-wise distance matrices are computed once in preprocessing step using NetworkX and Numpy packages. The shortest pair-wise distance matrix 
is computed by Floyd-Warshall algorithm, inversed in an element-wise fashion, and row-normalized using softmax.
\subsection{Hyper-Parameters}
In out experiments, we fixed $\alpha$=0.2 and $t$=5 for PPR and heat diffusion, respectively. We used grid search to choose the hyper-parameters from the following ranges. For graph classification, we chose the number of GCN layers, number of epochs, batch size, and the C parameter of the SVM from [2, 4, 8, 12], [10, 20, 40, 
100], [32, 64, 128, 256], and $ [10^{-3}$, $10^{-2}$, ..., $10^2$, $10^3$], respectively. For node classification, we set the number of GCN layers and the number of epochs and to 1 and 
2,000, respectively, and choose the batch size from [2, 4, 8]. We also use early stopping with a patience of 20. Finally, we set the size of hidden dimension of both 
node and graph representations to 512. The selected parameters are reported in Table \ref{table:appendix}.
\subsection{Effect of Number of Views}
Furthermore, we investigated whether increasing the number of views increases the performance on down-stream tasks, monotonically. We extended number of 
views to three by anchoring the main view on adjacency matrix and considering two diffusion matrices, i.e, PPR and heat, as other views. The results shown in 
Table \ref{tab:mv} suggest that unlike visual representation learning, extending the views does not improve the performance on the down-stream tasks. We 
speculate this is because different diffusion matrices carry similar information about the graph structure and hence using more of them does not introduce useful 
signals.
\subsection{Effect of Pooling}
In addition to mentioned sum pooling, we tried mean pooling and two variants of DiffPool: (1) DiffPool-1 where we use a single layer of DiffPool to aggregate all
node representations into a single graph representation, and (2) DiffPool-2 where we use two DiffPool layers to compute the intermediate representations. The first
layer projects nodes into a set of clusters, i.e., motifs, where the number of clusters is set as 25\% of the number of nodes before applying DiffPool, whereas the 
second layer projects the learned cluster representations into a single graph representations. It is noteworthy that to train the model with Diffpool layers, we jointly 
optimize the contrastive loss with an auxiliary link prediction loss and an entropy regularization. The results shown in Table \ref{table:appendix} suggest that sum 
pooling outperforms other pooling layers in 6 out of 8 benchmarks ,i.e., 5 out of 5 graph classification and 1 out if 3 node classification benchmarks. Also, results suggest that mean pooling achieves better results in 2 out of 3 node classification benchmarks.
\subsection{Effect of Encoder}
We also investigated the effect of assigning a dedicated encoder for each view or sharing an encoder across views. The results in Table \ref{table:appendix} show 
that using a dedicated encoder for each view consistently achieves better results across all benchmarks. When using shared encoder, we also randomly sampled 2 
out 4 views for each training sample in each mini-batch and contrasted the representation. We observed that unlike visual domain, it degraded the performance.
\subsection{Effect of Negative Samples}
Considering the importance of the number of negative samples in contrastive learning, we investigated the effect of batch size on the mode performance. For graph classification benchmarks, we used batch sizes of [16, 32, 64, 128, 256] and for classification benchmarks we evaluated batch sizes of [2, 4, 8]. As shown in Figure 2, we observe that increasing the batch size slightly increases the performance on graph classification task but has an negligible effect on the node classification.
\begin{figure}
\label{figbatch}
\vskip 0.2in
\begin{center}
\centerline{\includegraphics[width=170mm]{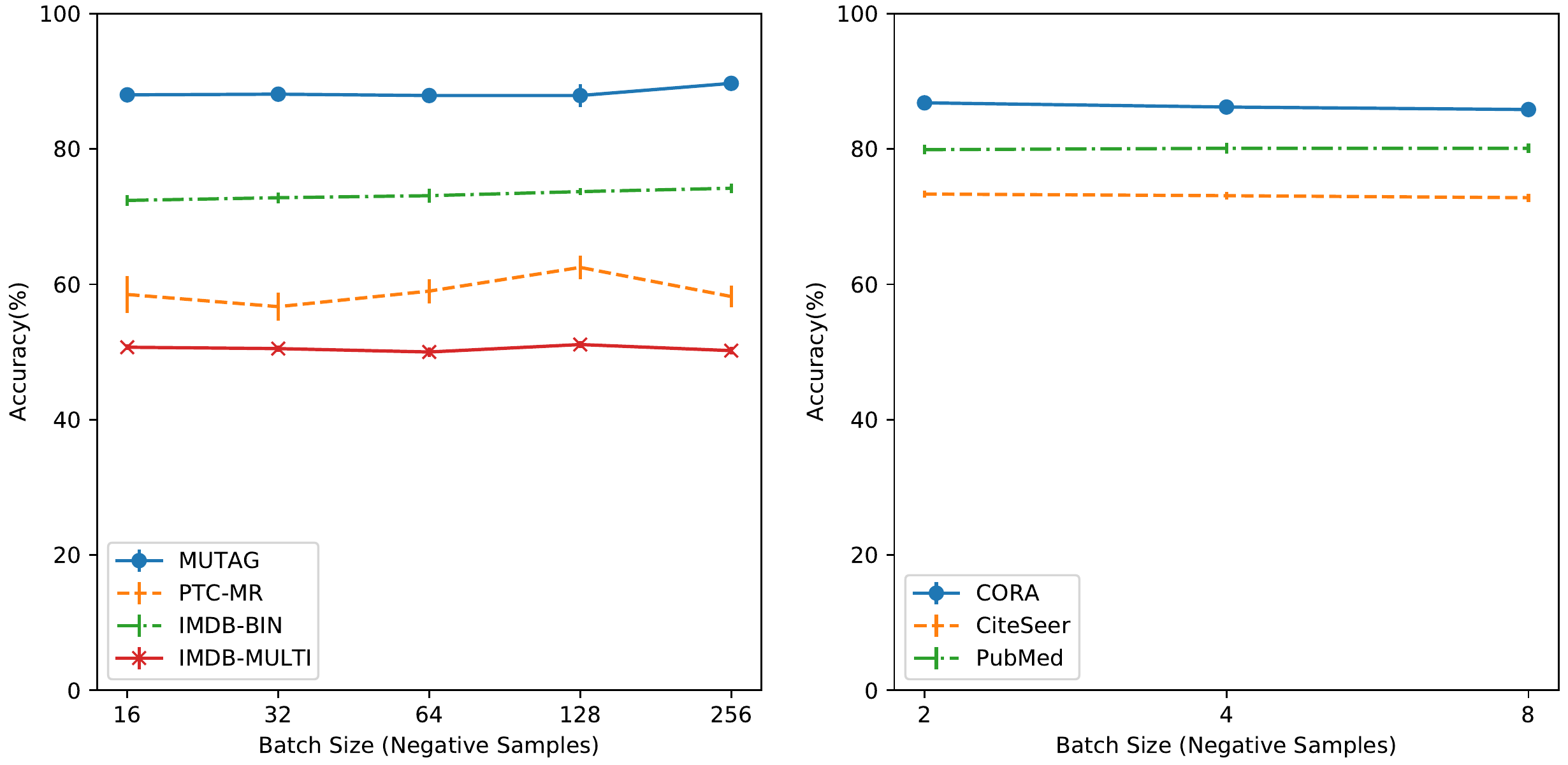}}
\caption{Effect of batch size (number of negative samples) on graph and node classification benchmarks. Increasing the batch size slightly increases the performance on graph classification task but has an negligible effect on the node classification.}
\end{center}
\vskip -0.2in
\end{figure}
\begin{table*}
\caption{Summary of chosen hyper-parameters using grid search, and effect of graph pooling and encoders on the accuracy on both node and graph classification benchmarks. Using sum pooling and assigning dedicated encoders for each view achieve better results across the benchmarks.}
\label{table:appendix}
\setlength{\tabcolsep}{3.5pt}
\vskip 0.2in
\begin{center}
\begin{small}
\begin{sc}
\begin{tabular}{clccc|ccccc}
\toprule
& ~ & \multicolumn{3}{c}{\textit{Node}} & \multicolumn{5}{c}{\textit{Graph}}\\
\cline{3-10}
& ~ & \textbf{cora} & \textbf{citeseer} & \textbf{pubmed} & \textbf{mutag} & \textbf{ptc-mr} & \textbf{imdb-bin} &\textbf{ imdb-multi} & \textbf{reddit-bin} \\
\midrule
\multirow{3}{*}{\begin{turn}{90}Hyper\end{turn}}
& $|$Layers$|$     & 1 & 1 & 1 & 4  &  4   &  2 & 4  & 2 \\
& $|$Batches$|$   & 2  & 2 & 2 & 20  &  40   &  20  &  40 & 20\\
& $|$Epochs$|$    & 2000  & 2000  & 2000 & 256  &  128 &  256  & 128  &  32 \\
\midrule
\multirow{4}{*}{\begin{turn}{90}Pooling\end{turn}}
& sum            & 86.2 $\pm$ 0.6 & \textbf{73.3 $\pm$ 0.5} & 79.6 $\pm$ 0.9 & \textbf{89.7 $\pm$ 1.1}  &  \textbf{62.5 $\pm$ 1.7}   &  \textbf{74.2 $\pm$ 0.7}  &  \textbf{51.1 $\pm$ 0.5}  &  \textbf{84.5 $\pm$ 0.6} \\
& mean         &  \textbf{86.8 $\pm$ 0.5}   & 73.2 $\pm$ 0.6   & \textbf{80.1 $\pm$ 0.7} & 88.8 $\pm$ 0.7  &  60.9 $\pm$ 1.2   &  72.3 $\pm$ 0.5  &  49.1 $\pm$ 0.7   & 79.4 $\pm$ 0.5 \\
& diffpool-1  & 84.6 $\pm$ 0.8 & 65.2  $\pm$ 1.7 & 76.1 $\pm$ 0.8 & 89.6  $\pm$ 0.9  &  61.1 $\pm$ 0.2  &  72.8 $\pm$ 0.6  &  49.4 $\pm$ 0.9   &  81.3 $\pm$ 0.4 \\
& diffpool-2  & 83.2 $\pm$ 0.9 & 63.5 $\pm$ 1.5 & 75.7 $\pm$ 1.1 & 88.0 $\pm$ 0.8  &  56.6 $\pm$ 1.8   &  72.7 $\pm$ 0.4  &  50.6 $\pm$ 0.5  &  82.8 $\pm$ 0.6 \\
\midrule
\multirow{2}{*}{\begin{turn}{90}Enc.\end{turn}}
& shared               & 86.2 $\pm$ 0.6 & 72.8 $\pm$ 0.6 & 79.5 $\pm$ 0.1 & 82.8 $\pm$ 1.9  &  55.9 $\pm$ 2.2   &  73.0 $\pm$ 0.7  &  50.0 $\pm$ 0.3  & 81.3 $\pm$ 2.0 \\
& dedicated          &\textbf{86.8 $\pm$ 0.5} & \textbf{73.3 $\pm$ 0.5} & \textbf{80.1 $\pm$ 0.7} & \textbf{89.7 $\pm$ 1.1}  & \textbf{62.5 $\pm$ 1.7}   &  \textbf{74.2 $\pm$ 0.7}  &  \textbf{51.1 $\pm$ 0.5}   & \textbf{84.5 $\pm$ 0.6} \\
\bottomrule
\end{tabular}
\end{sc}
\end{small}
\end{center}
\vskip -0.2in
\end{table*}
\begin{table}
\caption{Effect of number of views on node classification accuracy.}
\label{tab:mv}
\vskip 0.2in
\begin{center}
\begin{small}
\begin{sc}
\begin{tabular}{lccc}
\toprule
\textbf{\#Views} & \textbf{Cora} & \textbf{Citeseer} & \textbf{Pubmed} \\
\midrule
2 & \textbf{86.8 $\pm$ 0.5}   & \textbf{73.3 $\pm$ 0.5}   &\textbf{ 80.1 $\pm$ 0.7} \\
3 & 85.3 $\pm$ 0.5  & 71.2 $\pm$ 0.7 & 79.9 $\pm$ 0.6 \\
\bottomrule
\end{tabular}
\end{sc}
\end{small}
\end{center}
\vskip -0.1in
\end{table}

\end{document}